\documentclass[lettersize,journal]{IEEEtran}
\usepackage{amsmath,amsfonts}
\usepackage{xcolor}
\usepackage{algorithmic}
\usepackage{algorithm}
\usepackage{array}
\usepackage[caption=false,font=normalsize,labelfont=sf,textfont=sf]{subfig}
\usepackage{textcomp}
\usepackage{stfloats}
\usepackage{url}
\usepackage{verbatim}
\usepackage{graphicx}
\usepackage{enumitem}
\usepackage{listings}
\usepackage[backend=biber, style=ieee, url=false]{biblatex}

\begin{document}

\title{MAGIC: Transition-Aware Generation of Navigable Multi-Scene Game Worlds with Large Language Models}
\author{
\IEEEauthorblockN{Tsz Hei Fan\textsuperscript{*}, Choi Wing Fung\textsuperscript{*}, Yuxuan Wan\textsuperscript{\dag}, Shuqing Li\textsuperscript{\ddag}, and Michael R. Lyu}\\
\IEEEauthorblockA{Department of Computer Science and Engineering, The Chinese University of Hong Kong\\
\{thfan, serenefcw, yxwan\}@link.cuhk.edu.hk, \{sqli21, lyu\}@cse.cuhk.edu.hk}
\thanks{\textsuperscript{*} Fan and Fung contributed equally to this work and are co-first authors. \textsuperscript{\dag} Wan is the project lead. \textsuperscript{\ddag} Li is the corresponding author. This work has been submitted to the IEEE for possible publication. Copyright may be transferred without notice, after which this version may no longer be accessible.}
}




\maketitle

\begin{abstract}
Multi-scene navigation (clearing an objective in one bounded space and then crossing a portal into the next) is a defining feature of contemporary 3D games, but authoring it is laborious: every portal must have consistent endpoints on both sides, each interior must remain navigable once it is furnished, and the resulting connectivity must be kept consistent across many files. 
Recent large language model (LLM) and multimodal LLM (MLLM) scene generators have made single-interior synthesis dramatically cheaper, yet they produce one scene at a time and cannot, by naive repetition, yield a connected multi-scene world. 
We identify three obstacles that single-scene methods leave unsolved: cross-scene \emph{consistency}, in-scene \emph{navigability}, and the \emph{evaluation} of whether a transition actually works. 
We present MAGIC, a prompt-to-project system that addresses all three. 
MAGIC is a four-stage pipeline that turns a single natural-language prompt into a runnable multi-scene game project: it plans a shared transition-aware intermediate representation, specifies each scene while enforcing portal reachability with a flood-fill validator, generates the scenes together with their transition scripts, and combines them into one project. 
Because existing single-scene fidelity metrics never execute a transition, we further introduce a transition-focused evaluation agent that runs each transition in play. 
On a new benchmark of 100 multi-scene cases, MAGIC produces an executable project for every case and reaches 0.99 precision, 0.95 recall, and 0.96 F1 on end-to-end transition identification; stage by stage, it recovers more ground-truth portals and yields markedly more navigable layouts than an LLM baseline and Holodeck. Our code is available at \url{https://github.com/sereneee1201/MAGIC/}.
\end{abstract}

\begin{IEEEkeywords}
Multi-Scene Generation, Scene Transition, Game Development, Text-to-3D Scene, Large Language Model
\end{IEEEkeywords}

\section{Introduction}
Multi-scene navigation is a defining feature of contemporary 3D games. Dungeon crawlers, escape rooms, and RPGs all task the player with clearing an objective in one bounded space, then crossing a portal—a door, stairwell, or cutscene—into the next \cite{mti9060059}. For the designer, this involves two coupled problems: each scene must be a coherent, navigable interior, and every portal must have consistent endpoints on both sides matching position, destination, and transition effect. This process requires the designer to maintain a connection graph by hand and to keep paired portals, transition scripts, and doorway clearance consistent across files. Prior work on game development identifies this hand-authored connectivity as one of the most labor-intensive parts of 3D game production \cite{ghosh2022challenges}. 

Recent LLM- and MLLM-driven scene generators have made the first half of the problem (i.e., interior generation) dramatically cheaper, but the second half (i.e., scene transition and connectivity) remains challenging. Systems such as Holodeck \cite{yang2024holodeck} and Scenethesis \cite{Li20253DSS} synthesize a single furnished interior from a natural-language prompt, and surveys document a fast-growing text-to-3D landscape \cite{hong-etal-2025-game}.  However, none of the previous works generate inter-scene portals jointly with their destination scenes. Adjacent LLM-for-games work targets different artifacts (Sokoban-style 2D levels \cite{10.1145/3582437.3587211}, playable 2D environments from video \cite{10.5555/3692070.3692255}, single-environment narrative pipelines \cite{chen2025narrativetoscenegenerationllmdrivenpipeline}) rather than general 3D scenes.

Generating a connected multi-scene world exposes three obstacles that no naïve $n$-fold application of a single-scene model can resolve. \textbf{(C1) Consistency.} A transition is a \emph{joint} object: a portal is valid only if it exists on both sides with a consistent type and effect. Under stateless single-scene prompting, an MLLM tends to drop or hallucinate cross-scene references as the context grows, leaving the final multi-scene transition nonfunctional. \textbf{(C2) Navigability.} Each connected interior must remain \emph{navigable}: a portal can be named correctly on both sides yet be physically blocked once furniture is placed, and no single-scene metric flags this failure. \textbf{(C3) Evaluation.} No existing metric verifies whether a multi-scene transition actually works. Single-scene text-to-3D metrics measure visual fidelity or prompt alignment for one interior \cite{Wang2004ImageQA, Radford2021LearningTV}. They never execute a transition, so a mislinked portal or a broken transition script is invisible to them.

We present \textbf{MAGIC}, a \underline{\textbf{M}}ulti‑scene \underline{\textbf{A}}utomated \underline{\textbf{G}}ame worlds generator with \underline{\textbf{I}}ntelligent \underline{\textbf{C}}onnectivity. Figure~\ref {transition_pipeline} shows the overview of the system, which consists of four stages. The \emph{planning} stage produces per-scene descriptions plus a transition-aware intermediate representation (IR) shared across the pipeline, enforcing consistency~(C1). The \emph{scene specification} stage expands each scene and places objects, rejecting layouts whose portals fail a flood-fill reachability check on a 2D occupancy grid, enforcing navigability~(C2). The \emph{scene generation} stage materializes meshes and emits LevelLoader transition scripts bound to the correct portal colliders. The \emph{combination} stage stitches the per-scene Unity projects into one resulting project whose scenes reference each other via the emitted scripts. To address evaluation~(C3), we develop a transition-focused evaluation agent that operates directly on the packaged Unity project, executes each transition in play, and reports Precision, Recall, F1, Approach Rate, and Portal Match Rate against a ground-truth scene plan: a transition-aware complement to single-scene metrics. We evaluate MAGIC on a new benchmark of 100 multi-scene cases derived from MIT Indoor 67 \cite{5206537} and MMIS \cite{Kassab2024MMISMD}, comparing against an LLM baseline and Holodeck \cite{yang2024holodeck} stage by stage, plus an end-to-end evaluation of the full pipeline. MAGIC achieves 0.99 precision, 0.95 recall, and 0.96 F1 on end-to-end transition evaluation. This work contributes:
\begin{enumerate}[leftmargin=*]
\item An \textbf{end-to-end framework} that generates interconnected, navigable multi-scene game worlds from a single natural-language prompt, using a shared intermediate representation and a portal-reachability validator to enforce consistency and navigability across scenes.
\item A \textbf{multi-scene transition benchmark of 100 instances} covering scene graphs, looping and branching transition patterns, and mixed portal types. 
It is, to our knowledge, the first benchmark to provide ground-truth cross-scene transitions and portal reachability.
\item A \textbf{transition-focused evaluation agent} complementing single-scene text-to-3D metrics. The agent identifies candidate portals, executes each transition in play, navigates to each portal from the spawn point, applies an MLLM judge to assess portal appearance, and reports Precision, Recall, F1, Approach Rate, and Portal Match Rate against a ground-truth scene plan.
\item A \textbf{stage-by-stage and end-to-end evaluation} to establish the effectiveness of MAGIC across all pipeline components and demonstrate reliable transition generation from natural-language input alone.
\end{enumerate}

\section{Related Work}

\subsection{LLMs in Game Development}
Large Language Models (LLMs) have recently raised attention in game-related research, leading to a wide range of applications across different stages of game development. The roles of LLMs in game building span from runtime interaction to content creation. While early efforts have focused on conversational abilities and player interaction, more recent studies explore how LLMs can directly contribute to the design and generation of game components. Several works have categorized the roles of LLMs in games according to their function and timing within the development pipeline.

At runtime, LLMs have been widely used to create and control non‑player characters (NPCs), enabling dynamic dialogue and contextual responses during gameplay to enhance player experience \cite{li2025unbounded}. These works emphasize interaction and narrative engagement, leveraging natural language understanding to improve the player experience. In addition, recent research has shown that LLMs can contribute to the creation of game artifacts before the play begins. Acting as automated designers, LLMs can generate game content with minimal human contributions \cite{10645597}, they can also be the design assistants to support human designers in mixed‑initiative workflows, for example, translating natural‑language design intent into structured operations, or allowing interactive refinement within the unchanged domain constraints \cite{10645599}. LLMs have greatly enhanced the productivity of game development from an industry perspective by accelerating content creation \cite{Yermolaieva_2025}.

Studies on game level generation demonstrate that LLMs can produce structured environments \cite{10.1145/3582437.3587211}. By separating semantic reasoning from spatial realization, it is shown that more interpretable generation can be achieved while maintaining consistency with narrative intent \cite{chen2025narrativetoscenegenerationllmdrivenpipeline}. However, LLM-based gameworld generation has primarily focused on 2D environments and localized spatial content \cite{10.5555/3692070.3692255}, such as individual levels or standalone scenes. Explicit modeling of structural transitions and relationships between multiple game spaces remains underexplored.

\begin{table}[t]
    \caption{Comparison of different scene generation methods.}
    \centering
    \setlength{\tabcolsep}{3pt}
    \renewcommand{\arraystretch}{1.1}
    \begin{tabular}{lccccc}
        \hline
        Method & Text-Only & 3D Scene & Multi-Scene & Interactive \\
         & Input & Synthesis & Transition & Output \\
        \hline
        Genie \cite{10.5555/3692070.3692255} & \checkmark & $\times$ & $\times$ & \checkmark \\
        HOLODECK \cite{yang2024holodeck} & \checkmark & \checkmark & $\times$ & \checkmark \\
        Narrative-to-Scene \cite{chen2025narrativetoscenegenerationllmdrivenpipeline} & \checkmark & $\times$ & $\times$ & $\times$ \\
        OpenGame \cite{jiang2026opengameopenagenticcoding} & \checkmark & $\times$ & \checkmark & \checkmark \\
        UniGen \cite{yang202590} & \checkmark & \checkmark & $\times$ & \checkmark \\
        WonderTurbo \cite{Ni_2025_ICCV} & $\times$ & \checkmark & $\times$ & $\times$ \\
        WonderVerse \cite{feng2026wonderverseextendable3dscene} & \checkmark & \checkmark & $\times$ & $\times$ \\
        WonderWorld \cite{Yu_2025_CVPR} & $\times$ & \checkmark & $\times$ & $\times$ \\
        \hline
        \textbf{MAGIC (Ours)}
        & $\checkmark$ & $\checkmark$ & $\checkmark$ & $\checkmark$ \\
        \hline
    \end{tabular}
    \label{tab:comparison}
\end{table}

Table~\ref{tab:comparison} compares representative LLM- and MLLM-driven generators along four capabilities required by playable multi-scene games: a text-only prompt interface, genuine 3D scene synthesis, cross-scene transitions, and interactive output that drops into a game engine. The closest systems each cover only part of the problem. Holodeck~\cite{yang2024holodeck} and the Wonder family~\cite{Ni_2025_ICCV, feng2026wonderverseextendable3dscene, Yu_2025_CVPR} synthesize a single furnished or panoramic interior but never link scenes; Genie~\cite{10.5555/3692070.3692255} and Narrative-to-Scene~\cite{chen2025narrativetoscenegenerationllmdrivenpipeline} target 2D artifacts; and OpenGame~\cite{jiang2026opengameopenagenticcoding} produces interactive output without genuine 3D scene synthesis. None jointly generates a portal and its destination scene, so cross-scene transitions remain unaddressed. Producing a runnable Unity project, with transition scripts bound to colliders and a DSL-based layout, further frames multi-scene generation as a program-synthesis problem, connecting our pipeline to the constraint-expressive intermediate representation of Scenethesis~\cite{Li20253DSS}. MAGIC is the only method in Table~\ref{tab:comparison} that satisfies all four capabilities. 

\subsection{3D Scene Generation}
Researchers have conducted numerous studies on various 3D scene generation. Procedural content generation (PCG) remains an efficient method for automated scene creation by reducing development effort and cost \cite{maleki2024pcg}. It is particularly effective for generating repetitive and rule-based elements such as vegetation, buildings, and terrain \cite{Liu2024AnOO}. On the other hand, Neural 3D generation leverages models trained on 3D data to produce scene parameters such as object states and spatial relationships from text or image inputs \cite{Feng2025CasaGPTCA}. To address the limitations of 3D datasets, image-based approaches incorporate both images and text to enable photorealistic scene generation \cite{Rombach2021HighResolutionIS}, ensuring spatial consistency typically through panoramic inputs. Early methods applied GANs for image completion \cite{Akimoto2019360DegreeIC}, while more recent models improve controllability and output quality \cite{10044439}.

Recently, incorporating LLMs into the generation pipeline has stood out for its simplicity, requiring no strong mathematical or algorithmic expertise. Leveraging their strengths in linguistic and semantic comprehension \cite{Zeng2022SocraticMC}, in-context learning \cite{brown2020languagemodelsfewshotlearners}, and chain of thought \cite{wei2023chainofthoughtpromptingelicitsreasoning}, LLMs excel at understanding user intentions \cite{10.1145/3664647.3681129} and generating results that follow specific requirements.

Existing works show that the development of single-scene 3D generation has matured rapidly over the years, with active use of artificial intelligence. Therefore, our work focuses on developing a fully automated pipeline that establishes connections between scenes using action-triggered transitions, expanding the scope of scene-generation projects to multi-scene generation for 3D games.

\section{Methodology} \label{methodology}

\begin{figure*}[!t]
\centering
\includegraphics[width=\textwidth]{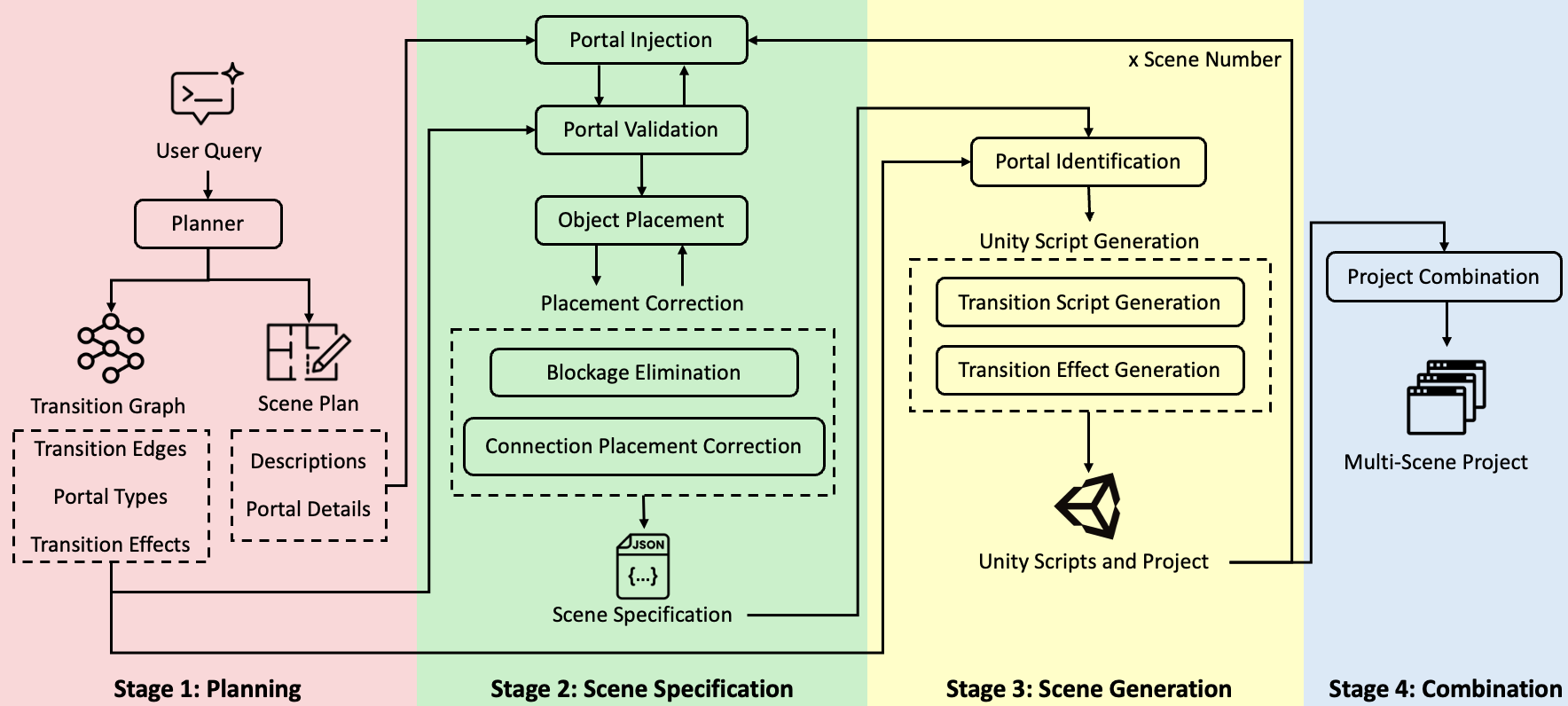}
\caption{The MAGIC pipeline and the elements that produce scene transitions, organized by the four stages: planning, scene specification, scene generation, and combination.}
\label{transition_pipeline}
\end{figure*}

MAGIC is a system that supports the generation of multiple playable scenes using a four-stage pipeline. It can automatically build customized connections between each scene for smooth navigation from a single input prompt, and finally merge the separate scenes into a complete multi-scene project.

The four stages are planning, scene specification, scene generation, and combination. The first stage accepts the user query as the only initial input. After analyzing and processing through a planner, the deliverables of this stage are a transition graph and multiple scene descriptions. Stage 2 and Stage 3 are repeated for each scene, taking each scene description from Stage 1 to output a corresponding scene specification. In Stage 3, the scene specification is assembled in Unity, a real-time 3D game development engine. The Unity projects for each scene are finally combined in Stage 4, producing a single multi-scene Unity project with the correct transitions.

\subsection{Stage 1: Planning} \label{stage-1}
The only input of this stage is the user query, which describes multiple scenes with transitions, and is processed by an LLM to produce individual scene prompts and a transition-aware automaton.

The LLM takes in a natural language input from the user and separates the prompt into individual scene descriptions while retaining all information from the original prompt. The LLM transforms user input into detailed and feasible scene descriptions as a professional game environment designer. Clear guidelines regarding the scene separation procedure are designed and provided to the LLM, with multiple sets of example input and output pairs.

To ensure that the styles of each scene are clearly differentiated, the LLM segregates the input into individual scenes to be processed independently in later stages. The scene descriptions are expanded to include 8 to 12 objects while maintaining internal consistency with the context. This style and object handling ensure that the scene generation model has abundant information to work with, regardless of the complexity of the initial prompt.

A transition-aware automaton records all connections between a pair of scenes with the portal and transition effects in a graph data structure. It is formulated as the following representation:
\begin{align*}
\label{pas-spec}
A = (S, P, \delta, E)
\end{align*}
where $S$ is the finite set of textual scene descriptions, $P$ is the finite set of portals, $E$ is the set of transition effects, and $\delta \subseteq S \times P \times E \times S$ is the transition relation: $(s_i, p, e, s_j) \in \delta$ means that portal $p$ connects scene $s_i$ to scene $s_j$ with effect $e$.

Because portals are traversable in both directions, $\delta$ is symmetric in its scene endpoints, $(s_i, p, e, s_j) \in \delta \Leftrightarrow (s_j, p, e, s_i) \in \delta$, so each transition is stored once as an undirected, effect-labeled edge. The LLM converts the extracted scene details into this representation, which doubles as a transition graph that downstream stages query and update; the graph is required to be connected so that every scene is reachable. To ensure that each required portal is mentioned in the scene descriptions, the number and type of portals required in the transition graph are counted and verified for each scene with a second LLM in a verification loop. Regarding the transition effects, ``FadeInOut" and ``IrisWipe" are currently available and will be chosen by the LLM if not specified. The decision is based on the stylistic and tonal difference in the neighboring scenes.

\subsection{Stage 2: Scene Specifications} \label{stage-2}
The algorithm takes in the scene descriptions and portal requirements and generates a complete and structured scene specification for Unity generation in Stage 3. Each scene is processed individually and iteratively to prevent the scenes from interfering with each other.

Before injecting the scene descriptions into the scene generation model, the transition-aware automaton is preprocessed to extract all the relevant portals for each scene. The portals are then sorted into connection portals and object portals to serve as validation requirements in the model. The object portals are assigned to different regions to prevent overlapping.

MAGIC first extracts regions from the scene prompt as subdivisions of the scene. Similar to Stage 1, it expands the scene descriptions with new objects and extended requirements. It then extracts sub-prompts from the newly expanded prompt by assigning snippets of the requirements to the regions, resulting in detailed region sub-prompts faithful to the original scene prompt. After region extractions, object injection is performed referencing the entity extraction method from Scenethesis \cite{Li20253DSS}. Each object attribute is filled in by the LLM according to the region prompt. The algorithm adopts a layered generation strategy as follows. It first requests a complete list of all required objects, then proceeds to fill in each layer of information across all objects before moving to the next. This ensures compatibility and style consistency across all objects and prevents global conflicts. 

Connections such as windows and doors are injected in the same way with the layered strategy, with the addition of a neighboring region pair. For the connection placements, they are initialized by the LLM but adjusted with a shifting algorithm. The original position generated by the LLM is often close to the wall but imprecise, as the model does not have a strong perception of 3D space. The algorithm first finds the shortest distance between the LLM placement of the connection and a wall, then determines the shifting direction according to the connection orientation. After attaching the connection to a wall, it is shifted inwards by half its thickness $T$ to make it more realistic.


To identify the portal among all the assets, the object portals are specified with an ``isPortal" tag in the object descriptions. It is added to the object injection stage to keep track of portal generation. To ensure that all object portals are generated, the verification process searches through objects with an ``isPortal" tag for the lists of expected object portals. With the list of missing portals passed to the LLM, the system regenerates a set of objects for each region until all object portals can be found in their assigned regions.

The connection portals are created in the connection injection stage. For doors, it is assumed that any doors connected to ``outside" are portals to another scene. The windows are only considered to be portals if their ``isPortal" attribute is true. The number of doors and windows is verified when the connections are named. The number of doors connected to the outside must match the expected amount so that there are no excess doors leading to nowhere unless specified otherwise. The ``isPortal" tags are validated later when the LLM fills in details for each connection. All the errors in door and window portal counts are passed into the LLM for regeneration, stating the number of respective portals expected to increase the chance of success in the following generation.

After retrieving all the objects, the LLM is asked to generate a set of positional and rotational constraints in a domain-specific language (DSL) developed in Scenethesis \cite{Li20253DSS} to list all object placement requirements. Based on the retrieved objects and requirements, the model will generate an initial placement as a draft version. The draft is then validated by the correction module, which compares the output with the placement constraints and lists all constraint violations for regeneration. This process is repeated until it has no violations or reaches the maximum iteration limit and returns the output with the highest constraint satisfaction rate.

To improve the player experience, a rigid body is attached to all objects so that the objects cannot be penetrated. This renders the elimination of the blockage crucial, as a scene transition is impossible if the doors are unreachable. A flood-fill algorithm is designed to detect blockage by rating the connectivity of each portal, which verifies whether every walkable cell in the scene is reachable from the portal.

First, the algorithm extracts the dimensions and positions of any obstacles and marks the space as a walkable area. The region is converted into a 2D occupancy grid:
\[
G \in \{0,1\}^{r \times c},
\]
\[
r = \left\lceil \frac{\max_z - \min_z}{s} \right\rceil, \quad
c = \left\lceil \frac{\max_x - \min_x}{s} \right\rceil,
\]
where $G_{ij} = 1$ indicates a walkable location, $G_{ij} = 0$ indicates an obstacle, and $s$ indicates the size of each cell, with $s = 0.05$ units in our experiments.

All objects are projected to the grid, and any cell covered is marked as blocked, except for carpet objects, which are placed flat on the floor, and ceiling-hung objects, which are placed in the ceiling and will not block the player's way to reach the transition portal. The flood-fill begins at the grid cell $g_p$ that corresponds to the position of the portal:
\[
\mathcal{V}_0 = \{\, g_p \,\}
\]
The reachable area is then expanded iteratively:
\[
\mathcal{V}_{k+1} =
\mathcal{V}_{k} \;\cup\; 
\Big\{ (i',j') \in \text{Adj}(i,j)
\;\Big|\;
(i,j) \in \mathcal{V}_k,\;\; G_{i'j'} = 1 \Big\}
\]
The final reachable set is:
\[
{
\mathcal{V} = \bigcup_{k=0}^{\infty} \mathcal{V}_k
}
\]

This process is repeated for all the portals in the region. A portal is considered to be reachable only if the number of walkable cells is equal to that of the visited cells. Once the algorithm has processed all portals in the region, it calculates the connectivity of the region as the fraction of walkable cells reached by the flood-fill:
\[
\mathrm{Connectivity} = \frac{|\mathcal{V}|}{\,|\{(i,j) : G_{ij}=1\}|\,}.
\]

If the ratio equals 1, all the portals in the region are considered to be reachable, and the region is fully navigable. A value lower than 1 indicates that there is a partial or complete blockage due to object placement. If the blockage evaluation is not successful, the object placement process will be looped until the maximum number of iterations is reached, and the model will return the placement with the highest connectivity score.

\subsection{Stage 3: Scene Generation} \label{stage-3}
After the complete scene specification is generated with scene details, region details, object details, and placements, it is passed to Stage 3 to generate assets that are necessary for the Unity project. The scene specification and assets are processed to produce an executable Unity project for a single scene, where the player can control the camera to navigate the scene. While the scripts for transitions are generated, they are not connected to any other scenes at this stage.

The scene specification is preprocessed to retrieve the portals generated by Stage 2. Any objects or connections with a true ``isPortal" tag and all doors connected to ``outside" are appended to the list of generated portals. The names of these objects are matched with the transition edges in the transition-aware automaton so that a list of transitions, each containing destination scene, portal name, and transition effect, is passed into the script generation system.

Meshes are generated by the composition algorithm in Scenethesis \cite{Li20253DSS} for all objects, walls, and floors. They are stored by their respective materials, vertices, and faces, while their paths are recorded in the scene specification file for easy referencing. The scripts are generated by processing information from the scene specification and filling in a pre-defined script template. This approach produces precise results and guarantees that the resulting scripts can be executed. 

As the addition of a rigid body to the camera object makes the triggering mechanism insensitive and unpredictable, the LevelLoader scripts for scene transition are attached to the portal objects instead when building the scene in Unity. The algorithm only needs to check if the object colliding with the portal is the camera object and load the correct scene with the required transition effect.

This algorithm first creates the folders necessary for a Unity project, then fills in the assets and scripts following the scene specification file generated in Stage 2. For asset models such as walls, objects, and floors, the system accesses the details from the specifications and converts them into a Unity-compatible format. It also retrieves the meshes generated in the previous stage and stores them in the correct Unity folder. The point lights are also injected into the folder according to the instructions in the specifications.

\subsection{Stage 4: Combination}
Stages 2 and 3 are repeated for each scene until they create the sets of outputs specified in the scene plan from Stage 1. Each output should contain the scene specification, object meshes, scene meshes, and a complete executable Unity file. Since the ultimate output only requires the Unity project, the combination algorithm discards excess information and merges the single-scene projects.

\section{Experiments and Evaluation}
The evaluation is divided into two parts, stage-by-stage and end-to-end, to specifically identify the performance of individual components of MAGIC and the overall performance for generating a multi-scene game environment. Regarding the first three stages of MAGIC, an LLM with a prompt describing the output format will serve as the baseline for all stages, while HOLODECK \cite{yang2024holodeck}, a single-scene generator that creates a scene specification before completing the software project, is an additional baseline for Stage 2. Since the final combination stage is a file management algorithm, it is not evaluated. In addition, the end-to-end evaluation is performed using a transition-focused evaluation agent, which we introduce for the first time to check whether the generated scenes are valid and contain all required transitions.


\subsection{Research Questions}
We organize the evaluation around the following five research questions:

\begin{itemize}
    \item \textbf{RQ1 (Planning):} Can Stage~1 recover an accurate per-scene plan and transition graph from a single multi-scene prompt?
    
    \item \textbf{RQ2 (Specification):} Are the portals produced in Stage~2 correct and physically reachable?
    
    \item \textbf{RQ3 (Generation):} Does Stage~3 yield executable Unity projects with the correct transition scripts?
    
    \item \textbf{RQ4 (Agent):} Does the transition-focused evaluation agent agree with human judgment?
    
    \item \textbf{RQ5 (End-to-end):} How accurately does the full pipeline generate functional cross-scene transitions?
\end{itemize}

\subsection{Experimental Setup}
For MAGIC, we used GPT-4.1-mini-2025-04-14 as the normal LLM. The LLM baseline used GPT-4.1. Unless otherwise specified by the underlying API, decoding followed the default settings, and the generation temperature was set to 0.7. All experiments were conducted on macOS and Windows, and a GPU is not required. The median duration was 33.35 minutes per scene (mean: 46.36 minutes), indicating a right-skewed distribution with a small number of long-running cases.

\subsection{Benchmark Test Cases} \label{test-cases}
To construct a diverse set of 100 test cases to evaluate the performance of MAGIC, two complementary datasets are utilized and combined to form the test cases, which are (1) the MIT 67 Indoor Scenes \cite{5206537} and (2) the Multimodal Interior Scenes (MMIS) \cite{Kassab2024MMISMD}.

The MIT 67 indoor scenes dataset provides a scene-centric database, which comprises a wide range of functional indoor categories, such as domestic environments, public spaces, and leisure venues. These varieties are designed to record large variability in spatial layout and object composition, making this dataset fit our assessment well. On the other hand, the MMIS dataset includes a large-scale collection of interior images, with textual annotations and audio descriptions across a range of different design styles. It decouples the room function from aesthetic style by representing multiple room types under diverse stylistic principles.

With a procedural script, 100 test cases are generated by integrating scene data from the two datasets. Each test case is a scene graph, where nodes are the scene instances and edges represent transitions between certain scene pairs. For each case, one to five scenes are selected from the MIT set, instantiated with characteristic objects and spatial constraints, and generated under a randomly sampled interior style from the MMIS taxonomy. The test cases are distributed across predefined patterns to ensure coverage of specific constraints, including but not limited to single-scene descriptions, looping transitions, and uniform or mixed portal types. Finally, each structured test case is serialized and passed to an LLM to generate a natural language prompt as input to the evaluation.

\subsection{Stage-by-Stage Evaluation}

\subsubsection{Scene Plan Evaluation}
The evaluation in Stage 1 aims to rate MAGIC's ability to convert a single multi-scene prompt into individual scene prompts and a transition-aware automaton. It is assessed by its description accuracy for each scene and graph accuracy for the transition-aware automaton.

The natural language prompt is processed by Stage 1 of our pipeline and the LLM baseline, which output a scene description for each scene and a transition-aware automaton detailing the origin, destination, portal type, and transition effect for each transition.

The scene description is evaluated by an LLM evaluator that extracts the requirement snippets from the ground truth dictionary and distributes them to each scene. The requirements are compared with the generated scene descriptions and return true if the requirement is mentioned. The scene accuracy is the fraction of ground-truth requirements that appear in the generated descriptions:
\[
\text{Scene Accuracy} = \frac{\displaystyle \sum_{i=1}^{N} R_i}{N}
\]

The transition-aware automaton is evaluated in three parts, namely the transition edges, portals, and effects. It is compared to the transition-aware automaton in the ground truth to output the accuracy score. The transition edges are considered present as long as the pair of neighbors is identical to that of the ground truth $GT\_edges$. The origin and destination are interchangeable as the transitions are bidirectional. Since missing edges risk disconnecting a scene, it has a greater weighting, showing its importance. The portal component compares the portals of the valid edges from the previous stage and divides the mismatched portals by the total number of portals in the ground truth. The effect component counts the number of valid effects and divides it by the total number of edges in the ground truth. The valid effects are ``FadeInOut'' and ``IrisWipe'' for processing in Stage 3 of the pipeline.

Scores for each aspect are summed to compute the final score of the transition-aware automaton. The portal accuracy is necessary for the Stage 2 and 3 of our pipeline to trigger the transition, while the effects are only utilized at Stage 3, thus the weighting of the portal score is higher than that of the effects.





\subsubsection{Scene Specification Evaluation}

Stage 2 evaluation focuses on the portals generated in the scene specifications and how they interact with other objects in the scenes. It compares the accuracy of the generated portals with the ground truth and calculates the blockage score for each portal per scene.

The input of Stage 2 is an accurate scene plan generated according to the ground truth. It is processed by Stage 2 of our pipeline, the HOLODECK baseline, and the LLM baseline.

The evaluator records the generated portals from the generated scene by finding the objects and windows with a true ``isPortal'' tag and appending all doors connected to ``outside''. The ground truth portals are extracted by locating transitions the scene is involved in and extracting their portals. The ground truth and generated portal lists are compared to find the number of matching, missing, and excess portals. Precision, recall, and F1 scores are calculated.

Blockage evaluation is conducted using the flood-fill algorithm introduced before. The algorithm measures whether each portal can be reached from all walkable cells in the region, thus rating the arrangement of the objects in terms of transition feasibility. Connectivity is calculated by dividing the number of visited cells by the walkable cells.

While a sparse region increases the ease of passing the flood-fill algorithm by having fewer obstacles, it also reduces the plausibility of the scene by the lack of objects. Thus the occupancy of the region is also calculated to measure its sparsity.

\subsubsection{Scene Generation}
Stage 3 evaluates how the program turns the scene specification into a playable single-scene Unity project. Since all assets are generated in Stage 2 of our pipeline, this stage focuses on script generation. It counts the number of LevelLoader scripts generated since it powers the transition process, and determines whether the generated scripts are executable.

The input is the scene plan used in the previous evaluation stage, and an accurate scene specification was generated from it. The inputs will go through a scene generation process identical to that of Stage 3 in our pipeline, except for the LLM baseline, where the script generation algorithm is replaced by an LLM.

The number of LevelLoaders generated in the resulting scripts should be equal to the number of transitions involving the scene in the transition-aware automaton of the scene plan. Missing LevelLoaders would mean unsuccessful transitions despite the portals being present, while excess portals will lead to unexpected transitions, thus both are undesirable. The scripts generated are evaluated by executing the SceneBuilder script, which will call the other generated scripts to create a full scene, and the execution will be considered unsuccessful if an error is raised at any point.

\subsection{End-to-End Evaluation}
\begin{figure*}[!t]
\centering
\includegraphics[width=\textwidth]{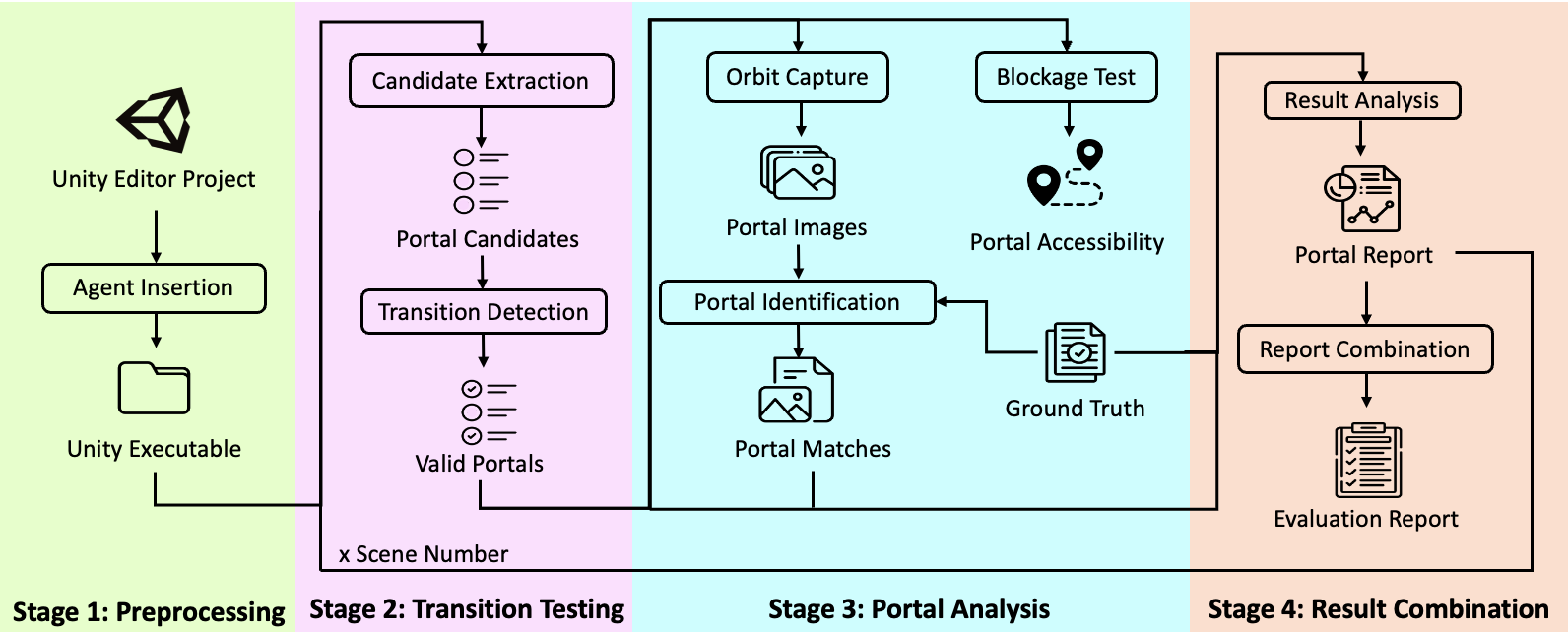}
\caption{The transition-focused evaluation agent. Operating on a packaged Unity project, it extracts portal candidates and tests each transition in play, analyzes portal accessibility and appearance with an MLLM judge against the ground truth, and aggregates the findings into an evaluation report.}
\label{eval_pipeline}
\end{figure*}

\subsubsection{Evaluation Agent Design}
Currently, there are several advanced methods and approaches to evaluate text-to-image or even text-to-3D generation results \cite{10.1145/3681758.3697992}. Unfortunately, while there are existing pixel-based metrics \cite{Wang2004ImageQA} and deep-learning-model-based metrics \cite{Radford2021LearningTV} that compare and compute similarity between the generation outputs with the reference images or the input prompts, there are rarely any methods proposed that focus on evaluating scene transitions. Noticing this gap in transition-focused evaluation tools, we constructed an automated evaluation agent to evaluate transition graph accuracy and portal appearance in actual gameplay. With reference to the report generated by this general agent, which can be applied to any Unity game project, game developers can review the outcome of their scene design and enhance the quality of the gaming environment.

The evaluation agent is inserted into the game project of each test case at the beginning and identifies possible transitions by screening objects in every game scene that could serve as scene transition candidates. These candidates are tested one by one by the agent to see if a scene transition occurs when the object is collided with, and it records the transited destination if so. For the portals that trigger transitions, the agent plays the role of a player and attempts to reach the portal from the spawning point, aiming to check its accessibility. Orbit captures are performed on the portals to record their appearance. The images are then sent to an MLLM to analyze whether the generated objects match the textual description from the original generation state. A final report is produced to summarize the overall scoring and metrics of the target game scene, comparing them with the criteria used for generation. 

\subsubsection{Reliability of the Evaluation Agent}
To evaluate the reliability of our evaluation agent's reports, 20 test cases are assessed and compared against human judgment ground truth. The test base is well-rounded, covering all scene counts, diverse interior room types and styles, and a variety of portal objects and transition edges.
To establish ground truth, every object in each test scene is inspected to determine whether it is a portal. If so, the portal will be manually activated, and the destination scene will be recorded. In addition, that object will also be checked to see if its appearance matches the expected portal shape and if it is accessible from the spawning position. All matching, missing, and extra transition edges are written in the human judgment report and are used to compare the performance of the evaluation agent. Two ablation studies are also designed and performed by removing portal candidate extraction and evaluating how its absence affects the agent's performance and efficiency, and examining the extent to which semantic cues available in the portal names, instead of captured images, are sufficient for evaluating portal correctness.

\begin{table}[ht]
\caption{Agreement of MAGIC's evaluation agent and two ablations (AB1, AB2) with human judgment, together with per-scene cost. For every metric except Duration, the entry is the \emph{signed difference} $\Delta=\text{human}-\text{method}$, and values closer to $0$ are better. MeanAbsDiff is the mean absolute difference across metrics. The best entry in each row is in \textbf{bold}.}
\centering
\setlength{\tabcolsep}{8pt}
\renewcommand{\arraystretch}{1.2}
\begin{tabular}{lccc}
\hline
 & AB1 & AB2 & MAGIC (Ours) \\
\hline
Duration per Scene (s) $\downarrow$ & $1081.61$ & $\mathbf{37.67}$ & $40.70$ \\
$\Delta$ Precision & $-0.5273$ & $\mathbf{0}$ & $\mathbf{0}$ \\
$\Delta$ Recall & $0.0146$ & $\mathbf{0}$ & $\mathbf{0}$ \\
$\Delta$ F1 & $-0.3737$ & $\mathbf{0}$ & $\mathbf{0}$ \\
$\Delta$ Approach Rate & $\mathbf{0.0009}$ & $0.0024$ & $0.0024$ \\
$\Delta$ Portal Match Rate & $\mathbf{-0.0448}$ & $0.1280$ & $0.0846$ \\
MeanAbsDiff & $0.2187$ & $0.0361$ & $\mathbf{0.0299}$ \\
\hline
\end{tabular}
\label{tab:agent-ablation}
\end{table}

Regarding the overall comparison, the results of the three mentioned methods are compared with those of human judgment. The differences between human judgment and the models are calculated by subtracting the method's value from the human judgment value, where a smaller difference indicates a higher degree of affinity, thus indicating greater evaluation accuracy. 

Our evaluation agent performs the best, with most metric values being superior to the others and the smallest difference from the human judgment results. Also, the mean absolute difference (MeanAbsDiff) between the evaluation agent and human judgment is the smallest. Since the only difference between our agent and Ablation Study 2 lies in the portal analysis section, while all other components remain unchanged, it is expected that the precision, recall, F1 score, and approach rate of these two methods are the same. In contrast, Ablation Study 1 includes many extra objects to be considered, which confuses the evaluation procedure, and even the low approach rate and portal rate are not reliable.

\section{Results and Analysis}

Our stage-by-stage and end-to-end evaluations target quantitative aspects; for qualitative aspects such as object placement and transition effects, we rely on visual analysis. In general, for cases that require the scenes to include doors or windows as the only type of portals, the results show that the connections are correctly placed on the walls without leaving any gaps. In terms of transition effects during the scene changes, both FadeInOut and IrisWipe are successfully implemented for the corresponding transitions. For other cases where the effects are not specified, IrisWipe is added for contrasting scenes, while FadeInOut is chosen for other transitions for gentle conversions. Regarding cases of varying prompt complexity, the outputs remain consistent, which shows that our pipeline can handle a range of prompt complexities.

\begin{figure}[!t]
\centering
\includegraphics[width=\columnwidth]{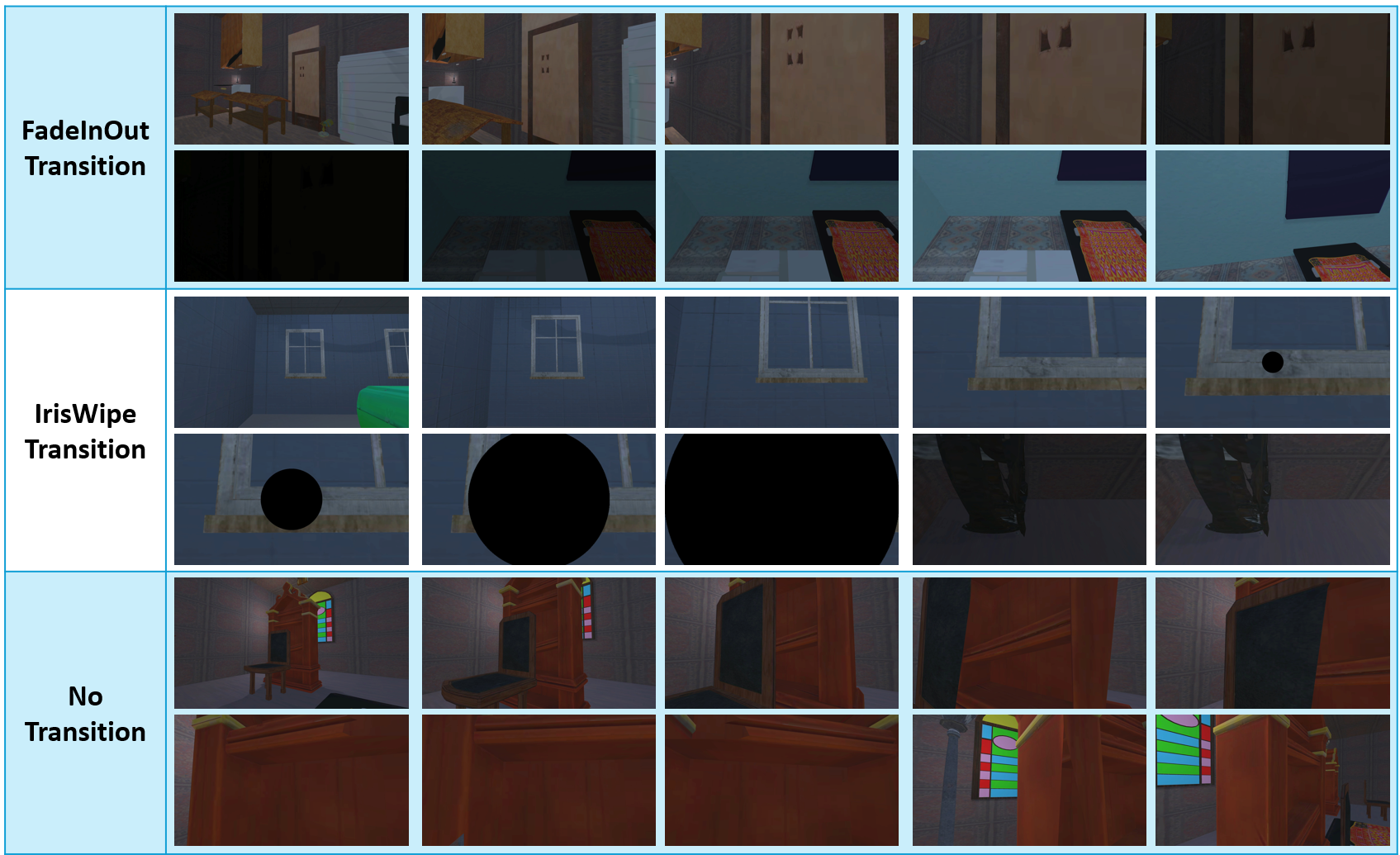}
\caption{Qualitative examples of transition effects in generated scenes, comparing FadeInOut, IrisWipe, and no transition across representative cases.}
\label{transition_pipeline}
\end{figure}

\subsection{Stage-by-Stage Evaluation Results}

\subsubsection{Scene Plan Evaluation Results}

Among all the test cases, the results of our own model win most of the cases as compared to the LLM baseline in terms of scene accuracy, as well as the transition-aware automaton. The result of having a relatively high accuracy score of both our own model and the LLM baseline is expected, as our model differs from the LLM baseline in more guidelines, and users' intentions are pre-defined in our Stage 1, yet still depends on the understanding ability of LLM models. What makes the largest difference is the transition-aware automaton due to the validation loops, which is our main contribution in this project.

\begin{table}[H]
\caption{Results of MAGIC and the LLM baseline after passing the planning stage. The best scores are in \textbf{BOLD}.}
\centering
\setlength{\tabcolsep}{8pt}
\renewcommand{\arraystretch}{1.2}
\begin{tabular}{lcc}
\hline
& LLM & \textbf{MAGIC (Ours)} \\
\hline
Average Scene Score & 0.9239 & \textbf{0.9672} \\
Average Graph Score & 0.9089 & \textbf{0.9711} \\
\hline
\end{tabular}
\end{table}

The consistent result in the evaluation of Stage 1 suggests that the planning stage of MAGIC can accurately identify the requirements and user intention through a single textual input. In addition, the results clearly demonstrated that the LLM baseline cannot maintain its quality for several scenarios, including those with transitions and multi-entity relationships, showing the significance of our planning stage in our end-to-end pipeline, as specified guidelines are necessary to be given to ensure the accurate interpretation of the user's thoughts.


\subsubsection{Scene Specification Evaluation Results}
MAGIC consistently achieves a comparatively large matched portion with the ground truth. LLM performs similarly, yet HOLODECK usually has fewer portals generated than required, suggesting its weaker understanding of transition logic. Regarding the missing counts, HOLODECK neglects the instructions frequently. In terms of missing counts, MAGIC performs the best.

\begin{table}[H]
\caption{Precision, recall, and F1 of the three methods. The best result in each row is in \textbf{bold}.}
\centering
\setlength{\tabcolsep}{8pt}
\renewcommand{\arraystretch}{1.2}
\begin{tabular}{lccc}
\hline
& LLM & HOLODECK & \textbf{MAGIC (Ours)} \\
\hline
Precision & \textbf{0.8825} & 0.7350 & 0.8333 \\
Recall & 0.9188 & 0.6026 & \textbf{0.9487} \\
F1 & \textbf{0.8863} & 0.6390 & 0.8709 \\
\hline
\end{tabular}
\end{table}

MAGIC stands out in the average connectivity score among all three methods. The nearly perfect value shows that portals generated by this method can achieve the almost complete reachability across walkable cells, preventing blockage, and ensuring the user experience during gameplay. In addition, its occupancy of nearly 30\% suggests that the scene generated with our method has a balanced density. The scene will not be too sparse or too compact. In comparison, although LLM ranked second in terms of its connectivity, its occupancy is the lowest among all methods. The third method, HOLODECK, results in the highest rate of occupancy, yet with the lowest rate of connectivity.

\begin{table}[H]
\caption{Connectivity and occupancy of the three methods. The best connectivity is in \textbf{bold}; occupancy is reported as a density indicator, where mid-range values (neither too sparse nor too dense) are preferable.}
\centering
\setlength{\tabcolsep}{8pt}
\renewcommand{\arraystretch}{1.2}
\begin{tabular}{lccc}
\hline
& LLM & HOLODECK & \textbf{MAGIC (Ours)} \\
\hline
Connectivity & 0.8718 & 0.8545 & \textbf{0.9952} \\
Occupancy & 0.1010 & 0.3971 & 0.2802 \\
\hline
\end{tabular}
\end{table}

Our method yields the highest recall, meaning it can identify nearly all the ground-truth portals with minimal misses. In addition, its precision is over 80\%, just slightly lower than that of LLM, and its F1 score is nearly 90\%, also ranked second among the three methods. Such a satisfactory performance shows that although our method sometimes generates extra portals, the overall balance between correctness and completeness is strong.

Alternatively, although the precision and F1 of LLM are the highest, its recall is lower than that of our proposed method, which means that LLM misses more ground-truth portals, and risks disconnecting scenes. HOLODECK performs poorly regarding the three metrics, which suggests that it faces misses and extras often, and is in line with its weaker connectivity.

\subsubsection{Scene Generation Results}
Regarding the success rate of creating a Unity project that can be executed, our model results in successes for all the test cases, whereas none of the projects generated directly by the LLM baseline can be executed. The reason is mainly that templates of a functional Unity project are not provided to the LLM baseline, thus its knowledge of the designer software is not sufficient to ensure all the essential components of an executable project are generated, even a minor error in the file naming can lead to an execution failure.

For most of the test cases, both MAGIC and the LLM baseline can generate the required number of LevelLoader. However, the LLM baseline generates excess LevelLoader scripts in nearly half of the test cases, which renders the transition logic unpredictable. Even if the Unity project generated by LLM can be executed, the transitions between the scenes must be implemented accurately for smooth gameplay.

\subsection{End-to-End Evaluation Results}
The performance of the generator is determined solely by the final multi-scene Unity project to produce an accurate transition evaluation. The evaluation agent tests the accuracy and quality of the transitions by deploying agents inside the Unity executable file to simulate in-game situations.

\begin{table}[H]
    \caption{Performance of MAGIC in the end-to-end evaluation.}
    \centering
    \setlength{\tabcolsep}{8pt}
    \renewcommand{\arraystretch}{1.2}
    \begin{tabular}{lc}
    \hline
    & \textbf{MAGIC (Ours)}\\
    \hline
    Precision & 0.9871\\
    Recall & 0.9481\\
    F1 & 0.9637\\
    Approach Rate & 0.9486\\
    Portal Match Rate & 0.7885\\
    \hline
    \end{tabular}
\end{table}

The accuracy of the generator is satisfactory as the precision, recall, and F1 score are all over 0.9, showing that most required portals are generated correctly. The portal match rate is lower than other metrics as the MLLM detection is strict and requires an exact match between the portal images and the ground truth, while the generation dataset might not cover all the portal types.

Most of the generated transitions align with the ground truth scene plan as precision, recall, and F1 score are all above 90\%. The precision score is narrower than recall, showing that the majority of the generated transitions correspond to portals in the ground truth, while some of them might be expected but are missing. The pipeline is thus proven to provide clean but less complete results, making it suitable for situations that require high transition precision.

We observe no clear correlation between the evaluation results and either the number of scenes or the expected portals per test case; that is, performance does not degrade with project scale over the tested range of one to five scenes. We therefore refrain from stronger scalability claims beyond this range. Also, it is shown that most portals can be reached from the project character spawn point, thus they are placed reasonably. The occurrence of inaccessible portals in a few cases is because the object placement module of the generator regenerates until the maximum number of retries is reached, where the placement with the highest connectivity is returned, and some blockages might still be present.

While the portal quality of MAGIC remains over 75\%, it is less satisfactory than other metrics, due to strict MLLM requirements and limited object mesh dataset. For example, one of the portal ground truths is ``directory board", and a plain board is generated as it is the most accurate mesh in the dataset. However, the MLLM requires evidence that the board is used for ``directory" purposes, thus the portal match fails. The strictness of the MLLM can be adjusted by modifying the system prompt to meet user expectations.

\subsection{Limitations}
MAGIC currently targets indoor scenes and the Unity engine, supports two transition effects (FadeInOut and IrisWipe), and accepts English text prompts only. Portal appearance is bounded by the coverage of the object-mesh dataset: when the closest available mesh lacks the semantic cue a portal name implies (e.g., a plain board for a ``directory board''), the MLLM judge correctly rejects it, which lowers the portal match rate without indicating a pipeline fault. Object placement is also best-effort: when the regeneration budget is exhausted, MAGIC returns the layout with the highest connectivity, so a small number of portals can remain blocked.

\subsection{Threats to Validity}
\textbf{Construct.} The ground-truth scene plans are generated procedurally, and metrics such as Approach Rate and Portal Match Rate are proxies for whether a human player can actually traverse the world; they may not capture every aspect of perceived transition quality. \textbf{Internal.} Both the pipeline and the baselines rely on stochastic LLMs, and the reported numbers come from a single run per case; without repeated runs and reported variance, point estimates may be optimistic, and we apply no significance testing. \textbf{External.} The benchmark comprises 100 synthetic indoor cases (one to five scenes) drawn from two datasets and is evaluated on a single engine and a single model family; generalization to outdoor or large-scale worlds, other engines, and other models is untested. \textbf{Conclusion.} The agent-reliability study uses only 20 cases and two human annotators reviewing the ground-truth transitions which limits the statistical strength of the agreement results.

\section{Conclusions}

MAGIC turns a single natural-language prompt into a runnable multi-scene 3D game project in which scenes are connected by consistent, navigable transitions. By sharing a transition-aware intermediate representation across the pipeline (C1), validating portal reachability with a flood-fill check on a 2D occupancy grid (C2), and introducing a transition-focused evaluation agent that executes each transition in play (C3), MAGIC addresses the three obstacles left open by single-scene generators. On a new benchmark of 100 multi-scene cases, it produces an executable project for every case and identifies cross-scene transitions with 0.99 precision, 0.95 recall, and 0.96 F1. Several directions remain. Multi-modal inputs and explicit portal-activation actions would broaden the range of expressible designs; human-in-the-loop intervention between stages (e.g., editing the scene specification before Stage~3) would raise the chance of a satisfactory result; and extending beyond indoor scenes, to additional transition effects, and to engines other than Unity would test how far the approach generalizes.


\printbibliography

@article{Li20253DSS,
    title={3D Software Synthesis Guided by Constraint-Expressive Intermediate Representation},
    author={Shuqing Li and Anson Y. Lam and Yun Peng and Wenxuan Wang and Michael R. Lyu},
    journal={ArXiv},
    year={2025},
    volume={abs/2507.18625},
    url={https://api.semanticscholar.org/CorpusID:280017510}
}

@article{Liu2024AnOO,
    title={An Overview of Procedurally Generating Virtual Environments},
    author={Yinghu Liu},
    journal={International Journal of Computer Science and Information Technology},
    year={2024},
    url={https://api.semanticscholar.org/CorpusID:268672916}
}

@article{Feng2025CasaGPTCA,
    title={CasaGPT: Cuboid Arrangement and Scene Assembly for Interior Design},
    author={Weitao Feng and Hang Zhou and Jing Liao and Li Cheng and Wenbo Zhou},
    journal={2025 IEEE/CVF Conference on Computer Vision and Pattern Recognition (CVPR)},
    year={2025},
    pages={29173-29182},
    url={https://api.semanticscholar.org/CorpusID:278165624}
}

@article{Rombach2021HighResolutionIS,
    title={High-Resolution Image Synthesis with Latent Diffusion Models},
    author={Robin Rombach and A. Blattmann and Dominik Lorenz and Patrick Esser and Bj{\"o}rn Ommer},
    journal={2022 IEEE/CVF Conference on Computer Vision and Pattern Recognition (CVPR)},
    year={2021},
    pages={10674-10685},
    url={https://api.semanticscholar.org/CorpusID:245335280}
}

@inproceedings{hong-etal-2025-game,
    title = "Game Development as Human-{LLM} Interaction",
    author = "Hong, Jiale and Wu, Hongqiu and Zhao, Hai",
    editor = "Che, Wanxiang and Nabende, Joyce and Shutova, Ekaterina and Pilehvar, Mohammad Taher",
    booktitle = "Proceedings of the 63rd Annual Meeting of the Association for Computational Linguistics (Volume 1: Long Papers)",
    month = jul,
    year = "2025",
    address = "Vienna, Austria",
    publisher = "Association for Computational Linguistics",
    url = "https://aclanthology.org/2025.acl-long.218/",
    doi = "10.18653/v1/2025.acl-long.218",
    pages = "4333--4354",
    ISBN = "979-8-89176-251-0",
}

@inproceedings{brown2020languagemodelsfewshotlearners,
    author = {Brown, Tom and Mann, Benjamin and Ryder, Nick and Subbiah, Melanie and Kaplan, Jared D and Dhariwal, Prafulla and Neelakantan, Arvind and Shyam, Pranav and Sastry, Girish and Askell, Amanda and Agarwal, Sandhini and Herbert-Voss, Ariel and Krueger, Gretchen and Henighan, Tom and Child, Rewon and Ramesh, Aditya and Ziegler, Daniel and Wu, Jeffrey and Winter, Clemens and Hesse, Chris and Chen, Mark and Sigler, Eric and Litwin, Mateusz and Gray, Scott and Chess, Benjamin and Clark, Jack and Berner, Christopher and McCandlish, Sam and Radford, Alec and Sutskever, Ilya and Amodei, Dario},
    booktitle = {Advances in Neural Information Processing Systems},
    pages = {1877--1901},
    title = {Language Models are Few-Shot Learners},
    url = {https://proceedings.neurips.cc/paper_files/paper/2020/file/1457c0d6bfcb4967418bfb8ac142f64a-Paper.pdf},
    volume = {33},
    year = {2020}
}

@article{wei2023chainofthoughtpromptingelicitsreasoning,
    author = {Wei, Jason and Wang, Xuezhi and Schuurmans, Dale and Bosma, Maarten and Ichter, Brian and Xia, Fei and Chi, Ed H. and Le, Quoc V. and Zhou, Denny},
    title = {Chain-of-Thought Prompting Elicits Reasoning in Large Language Models},
    year = {2022},
    journal = {Advances in Neural Information Processing Systems 35},
    numpages = {14}
}

@article{Zeng2022SocraticMC,
    title={Socratic Models: Composing Zero-Shot Multimodal Reasoning with Language},
    author={Andy Zeng and Adrian S. Wong and Stefan Welker and Krzysztof Choromanski and Federico Tombari and Aveek Purohit and Michael S. Ryoo and Vikas Sindhwani and Johnny Lee and Vincent Vanhoucke and Peter R. Florence},
    journal={ArXiv},
    year={2022},
    volume={abs/2204.00598},
    url={https://api.semanticscholar.org/CorpusID:247922520}
}

@inproceedings{10.1145/3664647.3681129,
    author = {Liu, Jia-Hong and Zhang, Shao-Kui and Zhang, Chuyue and Zhang, Song-Hai},
    title = {Controllable Procedural Generation of Landscapes},
    year = {2024},
    isbn = {9798400706868},
    publisher = {Association for Computing Machinery},
    address = {New York, NY, USA},
    url = {https://doi.org/10.1145/3664647.3681129},
    doi = {10.1145/3664647.3681129},
    booktitle = {Proceedings of the 32nd ACM International Conference on Multimedia},
    pages = {6394–6403},
    numpages = {10},
    location = {Melbourne VIC, Australia},
    series = {MM '24}
}

@article{ghosh2022challenges,
    author = {Ghosh, Aritra},
    year = {2022},
    month = {07},
    pages = {54-60},
    title = {Challenges in Educational Game Development},
    volume = {11},
    journal = {International Journal of Darshan Institute on Engineering Research \& Emerging Technology},
    doi = {10.32692/IJDI-ERET/11.1.2022.2209}
}

@Article{mti9060059,
    AUTHOR = {Swacha, Jakub},
    TITLE = {The Relative Popularity of Video Game Genres in the Scientific Literature: A Bibliographic Survey},
    JOURNAL = {Multimodal Technologies and Interaction},
    VOLUME = {9},
    YEAR = {2025},
    NUMBER = {6},
    ARTICLE-NUMBER = {59},
    URL = {https://www.mdpi.com/2414-4088/9/6/59},
    ISSN = {2414-4088},
    DOI = {10.3390/mti9060059}
}

@inproceedings{yang2024holodeck,
    title={Holodeck: Language guided generation of 3d embodied ai environments},
    author={Yang, Yue and Sun, Fan-Yun and Weihs, Luca and VanderBilt, Eli and Herrasti, Alvaro and Han, Winson and Wu, Jiajun and Haber, Nick and Krishna, Ranjay and Liu, Lingjie and others},
    booktitle={Proceedings of the IEEE/CVF Conference on Computer Vision and Pattern Recognition},
    pages={16227--16237},
    year={2024}
}

@inproceedings{maleki2024pcg,
    author = {Maleki, Mahdi Farrokhi and Zhao, Richard},
    title = {Procedural content generation in games: a survey with insights on emerging LLM integration},
    year = {2024},
    isbn = {1-57735-895-3},
    publisher = {AAAI Press},
    url = {https://doi.org/10.1609/aiide.v20i1.31877},
    doi = {10.1609/aiide.v20i1.31877},
    booktitle = {Proceedings of the Twentieth AAAI Conference on Artificial Intelligence and Interactive Digital Entertainment},
    articleno = {17},
    numpages = {12},
    location = {Lexington},
    series = {AIIDE '24}
}

@article{Akimoto2019360DegreeIC,
    title={360-Degree Image Completion by Two-Stage Conditional Gans},
    author={Naofumi Akimoto and Seito Kasai and Masaki Hayashi and Yoshimitsu Aoki},
    journal={2019 IEEE International Conference on Image Processing (ICIP)},
    year={2019},
    pages={4704-4708},
    url={https://api.semanticscholar.org/CorpusID:202763640}
}

@INPROCEEDINGS {10044439,
    author = { Dastjerdi, Mohammad Reza Karimi and Hold-Geoffroy, Yannick and Eisenmann, Jonathan and Khodadadeh, Siavash and Lalonde, Jean-Francois },
    booktitle = { 2022 International Conference on 3D Vision (3DV) },
    title = {Guided Co-Modulated GAN for 360° Field of View Extrapolation},
    year = {2022},
    volume = {},
    ISSN = {},
    pages = {475-485},
    doi = {10.1109/3DV57658.2022.00059},
    url = {https://doi.ieeecomputersociety.org/10.1109/3DV57658.2022.00059},
    publisher = {IEEE Computer Society},
    address = {Los Alamitos, CA, USA},
    month =sep
}

@article{Wang2004ImageQA,
    title={Image Quality Assessment: From Error Visibility to Structural Similarity},
    author={Zhou Wang and Alan Conrad Bovik and Hamid R. Sheikh and Eero P. Simoncelli},
    journal={IEEE Transactions on Image Processing},
    year={2004},
    volume={13},
    pages={600-612},
    url={https://api.semanticscholar.org/CorpusID:207761262}
}

@inproceedings{Radford2021LearningTV,
    title={Learning Transferable Visual Models From Natural Language Supervision},
    author={Alec Radford and Jong Wook Kim and Chris Hallacy and Aditya Ramesh and Gabriel Goh and Sandhini Agarwal and Girish Sastry and Amanda Askell and Pamela Mishkin and Jack Clark and Gretchen Krueger and Ilya Sutskever},
    booktitle={International Conference on Machine Learning},
    year={2021},
    url={https://api.semanticscholar.org/CorpusID:231591445}
}

@article{Kassab2024MMISMD,
    title={MMIS: Multimodal Dataset for Interior Scene Visual Generation and Recognition},
    author={Hozaifa Kassab and Ahmed Mahmoud and Mohamed Bahaa and Ammar Mohamed and Ali Hamdi},
    journal={2024 Intelligent Methods, Systems, and Applications (IMSA)},
    year={2024},
    pages={172-177},
    url={https://api.semanticscholar.org/CorpusID:271051085}
}

@INPROCEEDINGS{5206537,
    author={Quattoni, Ariadna and Torralba, Antonio},
    booktitle={2009 IEEE Conference on Computer Vision and Pattern Recognition}, 
    title={Recognizing Indoor Scenes}, 
    year={2009},
    volume={},
    number={},
    pages={413-420},
    doi={10.1109/CVPR.2009.5206537}
}

@inproceedings{10.1145/3681758.3697992,
    author = {Uchida, Yuiko and Togo, Ren and Maeda, Keisuke and Ogawa, Takahiro and Haseyama, Miki},
    title = {An Evaluation Metric for Single Image-to-3D Models Based on Object Detection Perspective},
    year = {2024},
    isbn = {9798400711404},
    publisher = {Association for Computing Machinery},
    address = {New York, NY, USA},
    url = {https://doi.org/10.1145/3681758.3697992},
    doi = {10.1145/3681758.3697992},
    booktitle = {SIGGRAPH Asia 2024 Technical Communications},
    articleno = {31},
    numpages = {4},
    location = {
    },
    series = {SA '24}
}

@inproceedings{li2025unbounded,
    title={Unbounded: A Generative Infinite Game of Character Life Simulation},
    author={Li, Jialu and Li, Yuanzhen and Wadhwa, Neal and Pritch, Yael and Jacobs, David E and Rubinstein, Michael and Bansal, Mohit and Ruiz, Nataniel},
    booktitle={International Conference on Learning Representations},
    volume={2025},
    pages={87465--87491},
    year={2025}
}

@INPROCEEDINGS{10645597,
    author={Hu, Chengpeng and Zhao, Yunlong and Liu, Jialin},
    booktitle={2024 IEEE Conference on Games (CoG)}, 
    title={Game Generation via Large Language Models}, 
    year={2024},
    volume={},
    number={},
    pages={1-4},
    doi={10.1109/CoG60054.2024.10645597}
}

@INPROCEEDINGS{10645599,
    author={Gallotta, Roberto and Liapis, Antonios and Yannakakis, Georgios},
    booktitle={2024 IEEE Conference on Games (CoG)}, 
    title={Consistent Game Content Creation via Function Calling for Large Language Models}, 
    year={2024},
    volume={},
    number={},
    pages={1-4},
    doi={10.1109/CoG60054.2024.10645599}
}

@article{Yermolaieva_2025,
    title={INTEGRATION OF LARGE LANGUAGE MODELS (LLM) INTO THE GAME CONTENT CREATION PROCESS},
    volume={4},
    url={https://sworldjournal.com/index.php/swj/article/view/swj30-04-075},
    DOI={10.30888/2663-5712.2025-30-04-075},
    number={30-04},
    journal={SWorldJournal},
    author={Yermolaieva, Yuliia},
    year={2025},
    month={Mar.},
    pages={74–83}
}

@inproceedings{10.1145/3582437.3587211,
    author = {Todd, Graham and Earle, Sam and Nasir, Muhammad Umair and Green, Michael Cerny and Togelius, Julian},
    title = {Level Generation Through Large Language Models},
    year = {2023},
    isbn = {9781450398558},
    publisher = {Association for Computing Machinery},
    address = {New York, NY, USA},
    url = {https://doi.org/10.1145/3582437.3587211},
    doi = {10.1145/3582437.3587211},
    booktitle = {Proceedings of the 18th International Conference on the Foundations of Digital Games},
    articleno = {70},
    numpages = {8},
    location = {Lisbon, Portugal},
    series = {FDG '23}
}

@article{chen2025narrativetoscenegenerationllmdrivenpipeline,
    title={Narrative-to-Scene Generation: An LLM-Driven Pipeline for 2D Game Environments},
    author={Chen, Yi-Chun and Jhala, Arnav},
    journal={arXiv preprint},
    year={2025}
}

@inproceedings{10.5555/3692070.3692255,
    author = {Bruce, Jake and Dennis, Michael and Edwards, Ashley and Parker-Holder, Jack and Shi, Yuge (Jimmy) and Hughes, Edward and Lai, Matthew and Mavalankar, Aditi and Steigerwald, Richie and Apps, Chris and Aytar, Yusuf and Bechtle, Sarah and Behbahani, Feryal and Chan, Stephanie and Heess, Nicolas and Gonzalez, Lucy and Osindero, Simon and Ozair, Sherjil and Reed, Scott and Zhang, Jingwei and Zolna, Konrad and Clune, Jeff and De Freitas, Nando and Singh, Satinder and Rockt\"{a}schel, Tim},
    title = {Genie: Generative Interactive Environments},
    year = {2024},
    publisher = {JMLR.org},
    booktitle = {Proceedings of the 41st International Conference on Machine Learning},
    articleno = {185},
    numpages = {21},
    location = {Vienna, Austria},
    series = {ICML'24}
}

@article{jiang2026opengameopenagenticcoding,
    title={OpenGame: Open Agentic Coding for Games}, 
    author={Yilei Jiang and Jinyuan Hu and Qianyin Xiao and Yaozhi Zheng and Ruize Ma and Kaituo Feng and Jiaming Han and Tianshuo Peng and Kaixuan Fan and Manyuan Zhang and Xiangyu Yue},
    year={2026},
    eprint={2604.18394},
    journal={ArXiv},
    primaryClass={cs.SE},
    url={https://arxiv.org/abs/2604.18394}, 
}

@article{yang202590,
    title={90\% Faster, 100\% Code-Free: MLLM-Driven Zero-Code 3D Game Development},
    author={Yang, Runxin and Wan, Yuxuan and Li, Shuqing and Lyu, Michael R},
    journal={arXiv},
    year={2025}
}

@InProceedings{Ni_2025_ICCV,
    author    = {Ni, Chaojun and Wang, Xiaofeng and Zhu, Zheng and Wang, Weijie and Li, Haoyun and Zhao, Guosheng and Li, Jie and Qin, Wenkang and Huang, Guan and Mei, Wenjun},
    title     = {WonderTurbo: Generating Interactive 3D World in 0.72 Seconds},
    booktitle = {Proceedings of the IEEE/CVF International Conference on Computer Vision (ICCV)},
    month     = {October},
    year      = {2025},
    pages     = {27423-27434}
}

@article{feng2026wonderverseextendable3dscene,
    title={WonderVerse: Extendable 3D Scene Generation with Video Generative Models}, 
    author={Hao Feng and Zhi Zuo and Jia-Hui Pan and Ka-Hei Hui and Qi Dou and Jingyu Hu and Zhengzhe Liu},
    year={2026},
    eprint={2503.09160},
    archivePrefix={arXiv},
    primaryClass={cs.CV},
    url={https://arxiv.org/abs/2503.09160}, 
    journal = {arXiv e-prints}
}

@InProceedings{Yu_2025_CVPR,
    author    = {Yu, Hong-Xing and Duan, Haoyi and Herrmann, Charles and Freeman, William T. and Wu, Jiajun},
    title     = {WonderWorld: Interactive 3D Scene Generation from a Single Image},
    booktitle = {Proceedings of the IEEE/CVF Conference on Computer Vision and Pattern Recognition (CVPR)},
    month     = {June},
    year      = {2025},
    pages     = {5916-5926}
}

\newpage

\begin{appendices}
\section{Evaluation Prompt for Stage 1 of MAGIC}

\begin{lstlisting}[basicstyle=\ttfamily\footnotesize, breaklines=true]
You are a professional environment art evaluator.

You will receive:
1. A dictionary containing the ground truth describing an interior scene.
2. A dictionary containing the generated scene description.

Your job:
- Output a list of scene names **EXACTLY** from the ground truth according to the indices in the generated scene description. 
- Evaluate how accurately the generated description match the ground truth.

**Accuracy Evaluation Rules:**
- Extract from the ground truth:
    - All key requirements that should appear in each scene.
        - Include scene types, styles, type of transition, positions of portals, object injections, and additional constraints.
        - For each transition, the type of transition **MUST** be considered a requirement for both scenes involved in the transition.
        - Do NOT consider the transition effects.
    - Any spatial or stylistic constraints (e.g., "table in the center", "warm lighting", "wooden furniture").
- For each extracted item:
    - If it appears and meets the constraint in generated description -> assign 1
    - If missing or violates constraint -> assign 0
- Compute success_rate for each scene:
    success_rate = (sum of successful checks) / (total required checks)
- Include a detailed checklist showing which items passed or failed.

**Output format (IMPORTANT):**
- Output must be valid JSON only - do NOT include Markdown fences, comments, or text outside the JSON object.
- Example:

Inputs:

Ground Truth:
{
  "case_6": {
    "number_of_scenes": 2,
    "details": {
      "1": {
        "scene_type": "office",
        "style": "industrial loft",
        "object injection": ["standing desk"],
        "additional constraints": [],
      },
      "2": {
        "scene_type": "storage room",
        "style": "functional",
        "object injection": ["steel shelving"],
        "additional constraints": ["exposed brick walls and steel beams"],
      }
    }
  },
  "transition": {
    "transition_graph": [
      { "from": "office", "to": "storage room", "type": "sliding metal partition", "effect": "IrisWipe", "position": "N/A"},
      { "from": "storage room", "to": "office", "type": "carpet", "effect": "N/A", "position": "N/A"},
    ]
  }
}


Generated Scene Description:
[
  {
    "scene_idx": 0,
    "prompt": "An office in an industrial loft style with exposed brick walls and steel beams. The room features a prominent standing desk made of reclaimed wood and metal. Large windows allow natural light to flood the space, casting shadows across the rugged wooden floor. Pipework is visible along the ceiling, adding to the industrial aesthetic. Near the back wall, a sliding metal partition serves as a gateway to the adjoining room."
  },
  {
    "scene_idx": 1,
    "prompt": "A functional-style storage room dominated by steel shelving units filled with neatly organized boxes and tools. The atmosphere is utilitarian but tidy, with a concrete floor and overhead LED strip lights providing crisp illumination. A carpet connects the storage room to the office, ensuring an easy and seamless transition between the spaces."
  }
]

Output:
{
  "nodes": ["office", "storage room"],
  "scenes":{
    "scene_1": {
        "accuracy": {
        "requirements": {
            "office": 1,
            "industrial loft": 1,
            "sliding metal partition": 1,
            "carpet": 0,
            "standing desk": 1
        },
        "success_rate": 0.8,
        }
    },
    "scene_2": {
        "accuracy": {
        "required_items": {
            "storage room": 1,
            "functional": 1,
            "sliding metal partition": 0,
            "carpet": 1,
            "exposed brick walls": 0,
            "steel beams": 0,
        },
        "success_rate": 0.5,
        }
    }
  }
}
\end{lstlisting}

\end{appendices}

\vfill

\end{document}